\def\a{\alpha}

\def\e{\epsilon}
\def\l{\lambda}

\def\r{\rightarrow}

\def\ie{{\it i.e.}}

\def\bi{\begin{itemize}}
\def\ei{\end{itemize}}

\newcounter{treecount}
\newcounter{branchcount}
\newsavebox{\parentbox}
\newsavebox{\treebox}
\newsavebox{\treeboxone}
\newsavebox{\treeboxtwo}
\newsavebox{\treeboxthree}
\newsavebox{\treeboxfour}
\newsavebox{\treeboxfive}
\newsavebox{\treeboxsix}
\newsavebox{\treeboxseven}
\newsavebox{\treeboxeight}
\newsavebox{\treeboxnine}
\newsavebox{\treeboxten}
\newsavebox{\treeboxeleven}
\newsavebox{\treeboxtwelve}
\newsavebox{\treeboxthirteen}
\newsavebox{\treeboxfourteen}
\newsavebox{\treeboxfifteen}
\newsavebox{\treeboxsixteen}
\newsavebox{\treeboxseventeen}
\newsavebox{\treeboxeighteen}
\newsavebox{\treeboxnineteen}
\newsavebox{\treeboxtwenty}
\newlength{\treeoffsetone}
\newlength{\treeoffsettwo}
\newlength{\treeoffsetthree}
\newlength{\treeoffsetfour}
\newlength{\treeoffsetfive}
\newlength{\treeoffsetsix}
\newlength{\treeoffsetseven}
\newlength{\treeoffseteight}
\newlength{\treeoffsetnine}
\newlength{\treeoffsetten}
\newlength{\treeoffseteleven}
\newlength{\treeoffsettwelve}
\newlength{\treeoffsetthirteen}
\newlength{\treeoffsetfourteen}
\newlength{\treeoffsetfifteen}
\newlength{\treeoffsetsixteen}
\newlength{\treeoffsetseventeen}
\newlength{\treeoffseteighteen}
\newlength{\treeoffsetnineteen}
\newlength{\treeoffsettwenty}

\newlength{\treeshiftone}
\newlength{\treeshifttwo}
\newlength{\treeshiftthree}
\newlength{\treeshiftfour}
\newlength{\treeshiftfive}
\newlength{\treeshiftsix}
\newlength{\treeshiftseven}
\newlength{\treeshifteight}
\newlength{\treeshiftnine}
\newlength{\treeshiftten}
\newlength{\treeshifteleven}
\newlength{\treeshifttwelve}
\newlength{\treeshiftthirteen}
\newlength{\treeshiftfourteen}
\newlength{\treeshiftfifteen}
\newlength{\treeshiftsixteen}
\newlength{\treeshiftseventeen}
\newlength{\treeshifteighteen}
\newlength{\treeshiftnineteen}
\newlength{\treeshifttwenty}
\newlength{\treewidthone}
\newlength{\treewidthtwo}
\newlength{\treewidththree}
\newlength{\treewidthfour}
\newlength{\treewidthfive}
\newlength{\treewidthsix}
\newlength{\treewidthseven}
\newlength{\treewidtheight}
\newlength{\treewidthnine}
\newlength{\treewidthten}
\newlength{\treewidtheleven}
\newlength{\treewidthtwelve}
\newlength{\treewidththirteen}
\newlength{\treewidthfourteen}
\newlength{\treewidthfifteen}
\newlength{\treewidthsixteen}
\newlength{\treewidthseventeen}
\newlength{\treewidtheighteen}
\newlength{\treewidthnineteen}
\newlength{\treewidthtwenty}
\newlength{\daughteroffsetone}
\newlength{\daughteroffsettwo}
\newlength{\daughteroffsetthree}
\newlength{\daughteroffsetfour}
\newlength{\branchwidthone}
\newlength{\branchwidthtwo}
\newlength{\branchwidththree}
\newlength{\branchwidthfour}
\newlength{\parentoffset}
\newlength{\treeoffset}
\newlength{\daughteroffset}
\newlength{\branchwidth}
\newlength{\parentwidth}
\newlength{\treewidth}
\newcommand{\ontop}[1]{\begin{tabular}{c}#1\end{tabular}}
\newcommand{\poptree}{%
\ifnum\value{treecount}=0\typeout{QobiTeX warning---Tree stack underflow}\fi%
\addtocounter{treecount}{-1}%
\setlength{\treeoffsettwo}{\treeoffsetthree}%
\setlength{\treeoffsetthree}{\treeoffsetfour}%
\setlength{\treeoffsetfour}{\treeoffsetfive}%
\setlength{\treeoffsetfive}{\treeoffsetsix}%
\setlength{\treeoffsetsix}{\treeoffsetseven}%
\setlength{\treeoffsetseven}{\treeoffseteight}%
\setlength{\treeoffseteight}{\treeoffsetnine}%
\setlength{\treeoffsetnine}{\treeoffsetten}%
\setlength{\treeoffsetten}{\treeoffseteleven}%
\setlength{\treeoffseteleven}{\treeoffsettwelve}%
\setlength{\treeoffsettwelve}{\treeoffsetthirteen}%
\setlength{\treeoffsetthirteen}{\treeoffsetfourteen}%
\setlength{\treeoffsetfourteen}{\treeoffsetfifteen}%
\setlength{\treeoffsetfifteen}{\treeoffsetsixteen}%
\setlength{\treeoffsetsixteen}{\treeoffsetseventeen}%
\setlength{\treeoffsetseventeen}{\treeoffseteighteen}%
\setlength{\treeoffseteighteen}{\treeoffsetnineteen}%
\setlength{\treeoffsetnineteen}{\treeoffsettwenty}%
\setlength{\treeshifttwo}{\treeshiftthree}%
\setlength{\treeshiftthree}{\treeshiftfour}%
\setlength{\treeshiftfour}{\treeshiftfive}%
\setlength{\treeshiftfive}{\treeshiftsix}%
\setlength{\treeshiftsix}{\treeshiftseven}%
\setlength{\treeshiftseven}{\treeshifteight}%
\setlength{\treeshifteight}{\treeshiftnine}%
\setlength{\treeshiftnine}{\treeshiftten}%
\setlength{\treeshiftten}{\treeshifteleven}%
\setlength{\treeshifteleven}{\treeshifttwelve}%
\setlength{\treeshifttwelve}{\treeshiftthirteen}%
\setlength{\treeshiftthirteen}{\treeshiftfourteen}%
\setlength{\treeshiftfourteen}{\treeshiftfifteen}%
\setlength{\treeshiftfifteen}{\treeshiftsixteen}%
\setlength{\treeshiftsixteen}{\treeshiftseventeen}%
\setlength{\treeshiftseventeen}{\treeshifteighteen}%
\setlength{\treeshifteighteen}{\treeshiftnineteen}%
\setlength{\treeshiftnineteen}{\treeshifttwenty}%
\setlength{\treewidthtwo}{\treewidththree}%
\setlength{\treewidththree}{\treewidthfour}%
\setlength{\treewidthfour}{\treewidthfive}%
\setlength{\treewidthfive}{\treewidthsix}%
\setlength{\treewidthsix}{\treewidthseven}%
\setlength{\treewidthseven}{\treewidtheight}%
\setlength{\treewidtheight}{\treewidthnine}%
\setlength{\treewidthnine}{\treewidthten}%
\setlength{\treewidthten}{\treewidtheleven}%
\setlength{\treewidtheleven}{\treewidthtwelve}%
\setlength{\treewidthtwelve}{\treewidththirteen}%
\setlength{\treewidththirteen}{\treewidthfourteen}%
\setlength{\treewidthfourteen}{\treewidthfifteen}%
\setlength{\treewidthfifteen}{\treewidthsixteen}%
\setlength{\treewidthsixteen}{\treewidthseventeen}%
\setlength{\treewidthseventeen}{\treewidtheighteen}%
\setlength{\treewidtheighteen}{\treewidthnineteen}%
\setlength{\treewidthnineteen}{\treewidthtwenty}%
\sbox{\treeboxtwo}{\usebox{\treeboxthree}}%
\sbox{\treeboxthree}{\usebox{\treeboxfour}}%
\sbox{\treeboxfour}{\usebox{\treeboxfive}}%
\sbox{\treeboxfive}{\usebox{\treeboxsix}}%
\sbox{\treeboxsix}{\usebox{\treeboxseven}}%
\sbox{\treeboxseven}{\usebox{\treeboxeight}}%
\sbox{\treeboxeight}{\usebox{\treeboxnine}}%
\sbox{\treeboxnine}{\usebox{\treeboxten}}%
\sbox{\treeboxten}{\usebox{\treeboxeleven}}%
\sbox{\treeboxeleven}{\usebox{\treeboxtwelve}}%
\sbox{\treeboxtwelve}{\usebox{\treeboxthirteen}}%
\sbox{\treeboxthirteen}{\usebox{\treeboxfourteen}}%
\sbox{\treeboxfourteen}{\usebox{\treeboxfifteen}}%
\sbox{\treeboxfifteen}{\usebox{\treeboxsixteen}}%
\sbox{\treeboxsixteen}{\usebox{\treeboxseventeen}}%
\sbox{\treeboxseventeen}{\usebox{\treeboxeighteen}}%
\sbox{\treeboxeighteen}{\usebox{\treeboxnineteen}}%
\sbox{\treeboxnineteen}{\usebox{\treeboxtwenty}}}
\newcommand{\leaf}[1]{%
\ifnum\value{treecount}=20\typeout{QobiTeX warning---Tree stack overflow}\fi%
\addtocounter{treecount}{1}%
\sbox{\treeboxtwenty}{\usebox{\treeboxnineteen}}%
\sbox{\treeboxnineteen}{\usebox{\treeboxeighteen}}%
\sbox{\treeboxeighteen}{\usebox{\treeboxseventeen}}%
\sbox{\treeboxseventeen}{\usebox{\treeboxsixteen}}%
\sbox{\treeboxsixteen}{\usebox{\treeboxfifteen}}%
\sbox{\treeboxfifteen}{\usebox{\treeboxfourteen}}%
\sbox{\treeboxfourteen}{\usebox{\treeboxthirteen}}%
\sbox{\treeboxthirteen}{\usebox{\treeboxtwelve}}%
\sbox{\treeboxtwelve}{\usebox{\treeboxeleven}}%
\sbox{\treeboxeleven}{\usebox{\treeboxten}}%
\sbox{\treeboxten}{\usebox{\treeboxnine}}%
\sbox{\treeboxnine}{\usebox{\treeboxeight}}%
\sbox{\treeboxeight}{\usebox{\treeboxseven}}%
\sbox{\treeboxseven}{\usebox{\treeboxsix}}%
\sbox{\treeboxsix}{\usebox{\treeboxfive}}%
\sbox{\treeboxfive}{\usebox{\treeboxfour}}%
\sbox{\treeboxfour}{\usebox{\treeboxthree}}%
\sbox{\treeboxthree}{\usebox{\treeboxtwo}}%
\sbox{\treeboxtwo}{\usebox{\treeboxone}}%
\sbox{\treeboxone}{\ontop{#1}}%
\sbox{\treeboxone}{\raisebox{-\ht\treeboxone}{\usebox{\treeboxone}}}%
\setlength{\treeoffsettwenty}{\treeoffsetnineteen}%
\setlength{\treeoffsetnineteen}{\treeoffseteighteen}%
\setlength{\treeoffseteighteen}{\treeoffsetseventeen}%
\setlength{\treeoffsetseventeen}{\treeoffsetsixteen}%
\setlength{\treeoffsetsixteen}{\treeoffsetfifteen}%
\setlength{\treeoffsetfifteen}{\treeoffsetfourteen}%
\setlength{\treeoffsetfourteen}{\treeoffsetthirteen}%
\setlength{\treeoffsetthirteen}{\treeoffsettwelve}%
\setlength{\treeoffsettwelve}{\treeoffseteleven}%
\setlength{\treeoffseteleven}{\treeoffsetten}%
\setlength{\treeoffsetten}{\treeoffsetnine}%
\setlength{\treeoffsetnine}{\treeoffseteight}%
\setlength{\treeoffseteight}{\treeoffsetseven}%
\setlength{\treeoffsetseven}{\treeoffsetsix}%
\setlength{\treeoffsetsix}{\treeoffsetfive}%
\setlength{\treeoffsetfive}{\treeoffsetfour}%
\setlength{\treeoffsetfour}{\treeoffsetthree}%
\setlength{\treeoffsetthree}{\treeoffsettwo}%
\setlength{\treeoffsettwo}{\treeoffsetone}%
\setlength{\treeoffsetone}{0.5\wd\treeboxone}%
\setlength{\treeshifttwenty}{\treeshiftnineteen}%
\setlength{\treeshiftnineteen}{\treeshifteighteen}%
\setlength{\treeshifteighteen}{\treeshiftseventeen}%
\setlength{\treeshiftseventeen}{\treeshiftsixteen}%
\setlength{\treeshiftsixteen}{\treeshiftfifteen}%
\setlength{\treeshiftfifteen}{\treeshiftfourteen}%
\setlength{\treeshiftfourteen}{\treeshiftthirteen}%
\setlength{\treeshiftthirteen}{\treeshifttwelve}%
\setlength{\treeshifttwelve}{\treeshifteleven}%
\setlength{\treeshifteleven}{\treeshiftten}%
\setlength{\treeshiftten}{\treeshiftnine}%
\setlength{\treeshiftnine}{\treeshifteight}%
\setlength{\treeshifteight}{\treeshiftseven}%
\setlength{\treeshiftseven}{\treeshiftsix}%
\setlength{\treeshiftsix}{\treeshiftfive}%
\setlength{\treeshiftfive}{\treeshiftfour}%
\setlength{\treeshiftfour}{\treeshiftthree}%
\setlength{\treeshiftthree}{\treeshifttwo}%
\setlength{\treeshifttwo}{\treeshiftone}%
\setlength{\treeshiftone}{0pt}%
\setlength{\treewidthtwenty}{\treewidthnineteen}%
\setlength{\treewidthnineteen}{\treewidtheighteen}%
\setlength{\treewidtheighteen}{\treewidthseventeen}%
\setlength{\treewidthseventeen}{\treewidthsixteen}%
\setlength{\treewidthsixteen}{\treewidthfifteen}%
\setlength{\treewidthfifteen}{\treewidthfourteen}%
\setlength{\treewidthfourteen}{\treewidththirteen}%
\setlength{\treewidththirteen}{\treewidthtwelve}%
\setlength{\treewidthtwelve}{\treewidtheleven}%
\setlength{\treewidtheleven}{\treewidthten}%
\setlength{\treewidthten}{\treewidthnine}%
\setlength{\treewidthnine}{\treewidtheight}%
\setlength{\treewidtheight}{\treewidthseven}%
\setlength{\treewidthseven}{\treewidthsix}%
\setlength{\treewidthsix}{\treewidthfive}%
\setlength{\treewidthfive}{\treewidthfour}%
\setlength{\treewidthfour}{\treewidththree}%
\setlength{\treewidththree}{\treewidthtwo}%
\setlength{\treewidthtwo}{\treewidthone}%
\setlength{\treewidthone}{\wd\treeboxone}}
\newcommand{\branch}[2]{%
\setcounter{branchcount}{#1}%
\ifnum\value{branchcount}=1\sbox{\parentbox}{\ontop{#2}}%
\setlength{\parentoffset}{\treeoffsetone}%
\addtolength{\parentoffset}{-0.5\wd\parentbox}%
\setlength{\daughteroffset}{0in}%
\ifdim\parentoffset<0in%
\setlength{\daughteroffset}{-\parentoffset}%
\setlength{\parentoffset}{0in}\fi%
\setlength{\parentwidth}{\parentoffset}%
\addtolength{\parentwidth}{\wd\parentbox}%
\setlength{\treeoffset}{\daughteroffset}%
\addtolength{\treeoffset}{\treeoffsetone}%
\setlength{\treewidth}{\wd\treeboxone}%
\addtolength{\treewidth}{\daughteroffset}%
\ifdim\treewidth<\parentwidth\setlength{\treewidth}{\parentwidth}\fi%
\sbox{\treebox}{\begin{minipage}{\treewidth}%
\begin{flushleft}%
\hspace*{\parentoffset}\usebox{\parentbox}\\
{\setlength{\unitlength}{2ex}%
\hspace*{\treeoffset}\begin{picture}(0,1)%
\put(0,0){\line(0,1){1}}%
\end{picture}}\\
\vspace{-\baselineskip}
\hspace*{\daughteroffset}%
\raisebox{-\ht\treeboxone}{\usebox{\treeboxone}}%
\end{flushleft}%
\end{minipage}}%
\setlength{\treeoffsetone}{\parentoffset}%
\addtolength{\treeoffsetone}{0.5\wd\parentbox}%
\setlength{\treeshiftone}{0pt}%
\setlength{\treewidthone}{\treewidth}%
\sbox{\treeboxone}{\usebox{\treebox}}%
\else\ifnum\value{branchcount}=2\sbox{\parentbox}{\ontop{#2}}%
\setlength{\branchwidthone}{\treewidthtwo}%
\addtolength{\branchwidthone}{\treeoffsetone}%
\addtolength{\branchwidthone}{-\treeshiftone}%
\addtolength{\branchwidthone}{-\treeoffsettwo}%
\setlength{\branchwidth}{\branchwidthone}%
\setlength{\daughteroffsetone}{\branchwidth}%
\addtolength{\daughteroffsetone}{-\branchwidthone}%
\addtolength{\daughteroffsetone}{-\treeshiftone}%
\setlength{\parentoffset}{-0.5\wd\parentbox}%
\addtolength{\parentoffset}{\treeoffsettwo}%
\addtolength{\parentoffset}{0.5\branchwidth}%
\setlength{\daughteroffset}{0in}%
\ifdim\parentoffset<0in%
\setlength{\daughteroffset}{-\parentoffset}%
\setlength{\parentoffset}{0in}\fi%
\setlength{\parentwidth}{\parentoffset}%
\addtolength{\parentwidth}{\wd\parentbox}%
\setlength{\treeoffset}{\daughteroffset}%
\addtolength{\treeoffset}{\treeoffsettwo}%
\setlength{\treewidth}{\wd\treeboxone}%
\addtolength{\treewidth}{\daughteroffsetone}%
\addtolength{\treewidth}{\treewidthtwo}%
\addtolength{\treewidth}{\daughteroffset}%
\ifdim\treewidth<\parentwidth\setlength{\treewidth}{\parentwidth}\fi%
\sbox{\treebox}{\begin{minipage}{\treewidth}%
\begin{flushleft}%
\hspace*{\parentoffset}\usebox{\parentbox}\\
{\setlength{\unitlength}{0.5\branchwidth}%
\hspace*{\treeoffset}\begin{picture}(2,0.5)%
\put(0,0){\line(2,1){1}}%
\put(2,0){\line(-2,1){1}}%
\end{picture}}\\
\vspace{-\baselineskip}
\hspace*{\daughteroffset}%
\makebox[\treewidthtwo][l]%
{\raisebox{-\ht\treeboxtwo}{\usebox{\treeboxtwo}}}%
\hspace*{\daughteroffsetone}%
\raisebox{-\ht\treeboxone}{\usebox{\treeboxone}}%
\end{flushleft}%
\end{minipage}}%
\setlength{\treeoffsetone}{\parentoffset}%
\addtolength{\treeoffsetone}{0.5\wd\parentbox}%
\setlength{\treeshiftone}{0pt}%
\setlength{\treewidthone}{\treewidth}%
\sbox{\treeboxone}{\usebox{\treebox}}\poptree%
\else\ifnum\value{branchcount}=3\sbox{\parentbox}{\ontop{#2}}%
\setlength{\branchwidthone}{\treewidthtwo}%
\addtolength{\branchwidthone}{\treeoffsetone}%
\addtolength{\branchwidthone}{-\treeshiftone}%
\addtolength{\branchwidthone}{-\treeoffsettwo}%
\setlength{\branchwidthtwo}{\treewidththree}%
\addtolength{\branchwidthtwo}{\treeoffsettwo}%
\addtolength{\branchwidthtwo}{-\treeshifttwo}%
\addtolength{\branchwidthtwo}{-\treeoffsetthree}%
\setlength{\branchwidth}{\branchwidthone}%
\ifdim\branchwidthtwo>\branchwidth%
\setlength{\branchwidth}{\branchwidthtwo}\fi%
\setlength{\daughteroffsetone}{\branchwidth}%
\addtolength{\daughteroffsetone}{-\branchwidthone}%
\addtolength{\daughteroffsetone}{-\treeshiftone}%
\setlength{\daughteroffsettwo}{\branchwidth}%
\addtolength{\daughteroffsettwo}{-\branchwidthtwo}%
\addtolength{\daughteroffsettwo}{-\treeshifttwo}%
\setlength{\parentoffset}{-0.5\wd\parentbox}%
\addtolength{\parentoffset}{\treeoffsetthree}%
\addtolength{\parentoffset}{\branchwidth}%
\setlength{\daughteroffset}{0in}%
\ifdim\parentoffset<0in%
\setlength{\daughteroffset}{-\parentoffset}%
\setlength{\parentoffset}{0in}\fi%
\setlength{\parentwidth}{\parentoffset}%
\addtolength{\parentwidth}{\wd\parentbox}%
\setlength{\treeoffset}{\daughteroffset}%
\addtolength{\treeoffset}{\treeoffsetthree}%
\setlength{\treewidth}{\wd\treeboxone}%
\addtolength{\treewidth}{\daughteroffsetone}%
\addtolength{\treewidth}{\treewidthtwo}%
\addtolength{\treewidth}{\daughteroffsettwo}%
\addtolength{\treewidth}{\treewidththree}%
\addtolength{\treewidth}{\daughteroffset}%
\ifdim\treewidth<\parentwidth\setlength{\treewidth}{\parentwidth}\fi%
\sbox{\treebox}{\begin{minipage}{\treewidth}%
\begin{flushleft}%
\hspace*{\parentoffset}\usebox{\parentbox}\\
{\setlength{\unitlength}{0.5\branchwidth}%
\hspace*{\treeoffset}\begin{picture}(4,1)%
\put(0,0){\line(2,1){2}}%
\put(2,0){\line(0,1){1}}%
\put(4,0){\line(-2,1){2}}%
\end{picture}}\\
\vspace{-\baselineskip}
\hspace*{\daughteroffset}%
\makebox[\treewidththree][l]%
{\raisebox{-\ht\treeboxthree}{\usebox{\treeboxthree}}}%
\hspace*{\daughteroffsettwo}%
\makebox[\treewidthtwo][l]%
{\raisebox{-\ht\treeboxtwo}{\usebox{\treeboxtwo}}}%
\hspace*{\daughteroffsetone}%
\raisebox{-\ht\treeboxone}{\usebox{\treeboxone}}%
\end{flushleft}%
\end{minipage}}%
\setlength{\treeoffsetone}{\parentoffset}%
\addtolength{\treeoffsetone}{0.5\wd\parentbox}%
\setlength{\treeshiftone}{0pt}%
\setlength{\treewidthone}{\treewidth}%
\sbox{\treeboxone}{\usebox{\treebox}}\poptree\poptree%
\else\ifnum\value{branchcount}=4\sbox{\parentbox}{\ontop{#2}}%
\setlength{\branchwidthone}{\treewidthtwo}%
\addtolength{\branchwidthone}{\treeoffsetone}%
\addtolength{\branchwidthone}{-\treeshiftone}%
\addtolength{\branchwidthone}{-\treeoffsettwo}%
\setlength{\branchwidthtwo}{\treewidththree}%
\addtolength{\branchwidthtwo}{\treeoffsettwo}%
\addtolength{\branchwidthtwo}{-\treeshifttwo}%
\addtolength{\branchwidthtwo}{-\treeoffsetthree}%
\setlength{\branchwidththree}{\treewidthfour}%
\addtolength{\branchwidththree}{\treeoffsetthree}%
\addtolength{\branchwidththree}{-\treeshiftthree}%
\addtolength{\branchwidththree}{-\treeoffsetfour}%
\setlength{\branchwidth}{\branchwidthone}%
\ifdim\branchwidthtwo>\branchwidth%
\setlength{\branchwidth}{\branchwidthtwo}\fi%
\ifdim\branchwidththree>\branchwidth%
\setlength{\branchwidth}{\branchwidththree}\fi%
\setlength{\daughteroffsetone}{\branchwidth}%
\addtolength{\daughteroffsetone}{-\branchwidthone}%
\addtolength{\daughteroffsetone}{-\treeshiftone}%
\setlength{\daughteroffsettwo}{\branchwidth}%
\addtolength{\daughteroffsettwo}{-\branchwidthtwo}%
\addtolength{\daughteroffsettwo}{-\treeshifttwo}%
\setlength{\daughteroffsetthree}{\branchwidth}%
\addtolength{\daughteroffsetthree}{-\branchwidththree}%
\addtolength{\daughteroffsetthree}{-\treeshiftthree}%
\setlength{\parentoffset}{-0.5\wd\parentbox}%
\addtolength{\parentoffset}{\treeoffsetfour}%
\addtolength{\parentoffset}{1.5\branchwidth}%
\setlength{\daughteroffset}{0in}%
\ifdim\parentoffset<0in%
\setlength{\daughteroffset}{-\parentoffset}%
\setlength{\parentoffset}{0in}\fi%
\setlength{\parentwidth}{\parentoffset}%
\addtolength{\parentwidth}{\wd\parentbox}%
\setlength{\treeoffset}{\daughteroffset}%
\addtolength{\treeoffset}{\treeoffsetfour}%
\setlength{\treewidth}{\wd\treeboxone}%
\addtolength{\treewidth}{\daughteroffsetone}%
\addtolength{\treewidth}{\treewidthtwo}%
\addtolength{\treewidth}{\daughteroffsettwo}%
\addtolength{\treewidth}{\treewidththree}%
\addtolength{\treewidth}{\daughteroffsetthree}%
\addtolength{\treewidth}{\treewidthfour}%
\addtolength{\treewidth}{\daughteroffset}%
\ifdim\treewidth<\parentwidth\setlength{\treewidth}{\parentwidth}\fi%
\sbox{\treebox}{\begin{minipage}{\treewidth}%
\begin{flushleft}%
\hspace*{\parentoffset}\usebox{\parentbox}\\
{\setlength{\unitlength}{0.5\branchwidth}%
\hspace*{\treeoffset}\begin{picture}(6,1)%
\put(0,0){\line(3,1){3}}%
\put(2,0){\line(1,1){1}}%
\put(4,0){\line(-1,1){1}}%
\put(6,0){\line(-3,1){3}}%
\end{picture}}\\
\vspace{-\baselineskip}
\hspace*{\daughteroffset}%
\makebox[\treewidthfour][l]%
{\raisebox{-\ht\treeboxfour}{\usebox{\treeboxfour}}}%
\hspace*{\daughteroffsetthree}%
\makebox[\treewidththree][l]%
{\raisebox{-\ht\treeboxthree}{\usebox{\treeboxthree}}}%
\hspace*{\daughteroffsettwo}%
\makebox[\treewidthtwo][l]%
{\raisebox{-\ht\treeboxtwo}{\usebox{\treeboxtwo}}}%
\hspace*{\daughteroffsetone}%
\raisebox{-\ht\treeboxone}{\usebox{\treeboxone}}%
\end{flushleft}%
\end{minipage}}%
\setlength{\treeoffsetone}{\parentoffset}%
\addtolength{\treeoffsetone}{0.5\wd\parentbox}%
\setlength{\treeshiftone}{0pt}%
\setlength{\treewidthone}{\treewidth}%
\sbox{\treeboxone}{\usebox{\treebox}}\poptree\poptree\poptree%
\else\ifnum\value{branchcount}=5\sbox{\parentbox}{\ontop{#2}}%
\setlength{\branchwidthone}{\treewidthtwo}%
\addtolength{\branchwidthone}{\treeoffsetone}%
\addtolength{\branchwidthone}{-\treeshiftone}%
\addtolength{\branchwidthone}{-\treeoffsettwo}%
\setlength{\branchwidthtwo}{\treewidththree}%
\addtolength{\branchwidthtwo}{\treeoffsettwo}%
\addtolength{\branchwidthtwo}{-\treeshifttwo}%
\addtolength{\branchwidthtwo}{-\treeoffsetthree}%
\setlength{\branchwidththree}{\treewidthfour}%
\addtolength{\branchwidththree}{\treeoffsetthree}%
\addtolength{\branchwidththree}{-\treeshiftthree}%
\addtolength{\branchwidththree}{-\treeoffsetfour}%
\setlength{\branchwidthfour}{\treewidthfive}%
\addtolength{\branchwidthfour}{\treeoffsetfour}%
\addtolength{\branchwidthfour}{-\treeshiftfour}%
\addtolength{\branchwidthfour}{-\treeoffsetfive}%
\setlength{\branchwidth}{\branchwidthone}%
\ifdim\branchwidthtwo>\branchwidth%
\setlength{\branchwidth}{\branchwidthtwo}\fi%
\ifdim\branchwidththree>\branchwidth%
\setlength{\branchwidth}{\branchwidththree}\fi%
\ifdim\branchwidthfour>\branchwidth%
\setlength{\branchwidth}{\branchwidthfour}\fi%
\setlength{\daughteroffsetone}{\branchwidth}%
\addtolength{\daughteroffsetone}{-\branchwidthone}%
\addtolength{\daughteroffsetone}{-\treeshiftone}%
\setlength{\daughteroffsettwo}{\branchwidth}%
\addtolength{\daughteroffsettwo}{-\branchwidthtwo}%
\addtolength{\daughteroffsettwo}{-\treeshifttwo}%
\setlength{\daughteroffsetthree}{\branchwidth}%
\addtolength{\daughteroffsetthree}{-\branchwidththree}%
\addtolength{\daughteroffsetthree}{-\treeshiftthree}%
\setlength{\daughteroffsetfour}{\branchwidth}%
\addtolength{\daughteroffsetfour}{-\branchwidthfour}%
\addtolength{\daughteroffsetfour}{-\treeshiftfour}%
\setlength{\parentoffset}{-0.5\wd\parentbox}%
\addtolength{\parentoffset}{\treeoffsetfive}%
\addtolength{\parentoffset}{2\branchwidth}%
\setlength{\daughteroffset}{0in}%
\ifdim\parentoffset<0in%
\setlength{\daughteroffset}{-\parentoffset}%
\setlength{\parentoffset}{0in}\fi%
\setlength{\parentwidth}{\parentoffset}%
\addtolength{\parentwidth}{\wd\parentbox}%
\setlength{\treeoffset}{\daughteroffset}%
\addtolength{\treeoffset}{\treeoffsetfive}%
\setlength{\treewidth}{\wd\treeboxone}%
\addtolength{\treewidth}{\daughteroffsetone}%
\addtolength{\treewidth}{\treewidthtwo}%
\addtolength{\treewidth}{\daughteroffsettwo}%
\addtolength{\treewidth}{\treewidththree}%
\addtolength{\treewidth}{\daughteroffsetthree}%
\addtolength{\treewidth}{\treewidthfour}%
\addtolength{\treewidth}{\daughteroffsetfour}%
\addtolength{\treewidth}{\treewidthfive}%
\addtolength{\treewidth}{\daughteroffset}%
\ifdim\treewidth<\parentwidth\setlength{\treewidth}{\parentwidth}\fi%
\sbox{\treebox}{\begin{minipage}{\treewidth}%
\begin{flushleft}%
\hspace*{\parentoffset}\usebox{\parentbox}\\
{\setlength{\unitlength}{0.5\branchwidth}%
\hspace*{\treeoffset}\begin{picture}(8,1)%
\put(0,0){\line(4,1){4}}%
\put(2,0){\line(2,1){2}}%
\put(4,0){\line(0,1){1}}%
\put(6,0){\line(-2,1){2}}%
\put(8,0){\line(-4,1){4}}%
\end{picture}}\\
\vspace{-\baselineskip}
\hspace*{\daughteroffset}%
\makebox[\treewidthfive][l]%
{\raisebox{-\ht\treeboxfour}{\usebox{\treeboxfive}}}%
\hspace*{\daughteroffsetfour}%
\makebox[\treewidthfour][l]%
{\raisebox{-\ht\treeboxfour}{\usebox{\treeboxfour}}}%
\hspace*{\daughteroffsetthree}%
\makebox[\treewidththree][l]%
{\raisebox{-\ht\treeboxthree}{\usebox{\treeboxthree}}}%
\hspace*{\daughteroffsettwo}%
\makebox[\treewidthtwo][l]%
{\raisebox{-\ht\treeboxtwo}{\usebox{\treeboxtwo}}}%
\hspace*{\daughteroffsetone}%
\raisebox{-\ht\treeboxone}{\usebox{\treeboxone}}%
\end{flushleft}%
\end{minipage}}%
\setlength{\treeoffsetone}{\parentoffset}%
\addtolength{\treeoffsetone}{0.5\wd\parentbox}%
\setlength{\treeshiftone}{0pt}%
\setlength{\treewidthone}{\treewidth}%
\sbox{\treeboxone}{\usebox{\treebox}}\poptree\poptree\poptree\poptree%
\else\typeout{QobiTeX warning--- Can't handle #1 branching}\fi\fi\fi\fi\fi}
\newcommand{\tree}{%
\usebox{\treeboxone}
\setlength{\treeoffsetone}{\treeoffsettwo}%
\sbox{\treeboxone}{\usebox{\treeboxtwo}}%
\poptree}

\font\tenmsb=msbm10
\font\sevenmsb=msbm7
\font\fivemsb=msbm5
\newfam\msbfam
\textfont\msbfam=\tenmsb
\scriptfont\msbfam=\sevenmsb
\scriptscriptfont\msbfam=\fivemsb
\def\Bbb#1{\fam\msbfam\relax#1}

\documentstyle[aclap]{article}

\title{\vspace{-0.5in}Bayesian Grammar Induction for Language Modeling}

\author{Stanley F. Chen \\
Aiken Computation Laboratory \\
Division of Applied Sciences \\
Harvard University \\
Cambridge, MA 02138 \\
{\tt sfc@das.harvard.edu}}

\begin{document}

\maketitle

\begin{abstract}
We describe a corpus-based induction algorithm for
probabilistic context-free grammars.  The algorithm employs a greedy
heuristic search within a Bayesian framework, and a post-pass
using the Inside-Outside algorithm.  We compare the performance
of our algorithm to $n$-gram models and the Inside-Outside algorithm
in three language modeling tasks.  In two of the tasks, the training
data is generated by a probabilistic context-free grammar and in both tasks
our algorithm outperforms the other techniques.  The third task involves
naturally-occurring data, and in this task our algorithm does not perform
as well as $n$-gram models but vastly outperforms the Inside-Outside algorithm.
\end{abstract}


\section{Introduction}

In applications such as speech recognition, handwriting recognition,
and spelling correction,
performance is limited by the quality of the language model utilized
\cite{Bahl:78a,Baker:75a,Kernighan:90a,Srihari:92a}.  However,
static language modeling performance has remained basically unchanged
since the advent of $n$-gram language models forty years ago
\cite{Shannon:51a}.  Yet, $n$-gram language models can only
capture dependencies within
an $n$-word window, where currently the largest practical $n$ for
natural language is three, and many dependencies in natural
language occur beyond a three-word window.  In addition, $n$-gram
models are extremely large, thus making them difficult
to implement efficiently in memory-constrained applications.

An appealing alternative is grammar-based language models.  Language
models expressed as a probabilistic grammar tend to be more compact than
$n$-gram language models, and have the ability to model long-distance
dependencies \cite{Lari:90a,Resnik:92a,Schabes:92a}.  However, to date
there has been little success in constructing grammar-based language
models competitive with $n$-gram models in problems of any
magnitude.

In this paper, we describe a corpus-based induction algorithm for
probabilistic context-free grammars that outperforms $n$-gram
models and the Inside-Outside algorithm \cite{Baker:79b}
in medium-sized domains.  This result marks the first
time a grammar-based language model has surpassed $n$-gram modeling
in a task of at least moderate size.  The algorithm employs a greedy
heuristic search within a Bayesian framework, and a post-pass
using the Inside-Outside algorithm.


\begin{table*}[tb]

$$\begin{array}{l}
\begin{array}{ccllcl}
S & \r & SX \hspace{0.3in} & (1 - \e) \\
S & \r & X  & (\e) \\
X & \r & A & (p(A)) & \;\;\;\;\; & \forall \; A \in N - \{S, X\} \\
A_a & \r & a & (1) & & \forall \; a \in T
\end{array} \\
\\
\begin{array}{ccl}
N & = & \mbox{the set of all nonterminal symbols} \\
T & = & \mbox{the set of all terminal symbols} \\
\\
\multicolumn{3}{l}{\mbox{Probabilities for each rule are in parentheses.}}
\end{array}
\end{array} $$
\caption{Initial hypothesis grammar \label{tab:init}}
\end{table*}

\section{Grammar Induction as Search}

Grammar induction can be framed as a search problem, and has been framed as
such almost without exception in past research \cite{Angluin:83a}.
The search space is taken to be some class of grammars; for example,
in our work we search within the space of probabilistic context-free
grammars.  The objective function is taken to be
some measure dependent on the training data; one generally wants
to find a grammar that in some sense accurately models the training data.

Most work in language modeling, including $n$-gram models and the
Inside-Outside algorithm, falls under the maximum-likelihood paradigm,
where one takes the objective function to be the likelihood
of the training data given the grammar.  However, the optimal grammar under
this objective function is one which generates
only strings in the training data and no other strings.  Such grammars
are poor language models, as they {\it overfit\/} the training data and
do not model the language at large.  In $n$-gram models and the
Inside-Outside algorithm, this issue is evaded by bounding the
size and form of the grammars considered, so that the
``optimal'' grammar cannot be expressed.  However, in our
work we do not wish to limit the size of the grammars considered.

The basic shortcoming of the maximum-likelihood objective function is that
it does not encompass the compelling intuition
behind Occam's Razor, that simpler (or smaller) grammars are preferable over
complex (or larger) grammars.  A factor in the objective function
that favors smaller grammars over large can prevent the objective function from
preferring grammars that overfit the training data.
\newcite{Solomonoff:64a} presents a Bayesian grammar induction framework
that includes such a factor in a motivated manner.

The goal of grammar induction is taken to be finding the grammar with
the largest {\it a posteriori\/} probability given the training data,
that is, finding the grammar $G'$ where
$$ G' = \arg\max_G p(G | O) $$
and where we denote the training data as $O$, for {\it observations}.
As it is unclear how to estimate $p(G|O)$ directly, we apply
Bayes' Rule and get
$$ G' = \arg\max_G \frac{p(O|G)p(G)}{p(O)} = \arg\max_G p(O|G)p(G) $$
Hence, we can frame the search for $G'$ as a search with the objective function
$p(O|G) p(G)$, the likelihood of the training data multiplied by the
prior probability of the grammar.

We satisfy the goal of favoring smaller grammars by choosing a prior
that assigns higher probabilities to such grammars.  In particular,
Solomonoff proposes the use of the {\it universal a priori probability}
\cite{Solomonoff:60a}, which is closely related to the
minimum description length principle later proposed by \cite{Rissanen:78a}.
In the case of grammatical language modeling, this
corresponds to taking
$$ p(G) = 2^{-l(G)} $$
where $l(G)$ is the length of the description of the grammar
in bits.  The universal {\it a priori\/} probability has many elegant
properties, the most salient of which is that it dominates all other
enumerable probability distributions multiplicatively.\footnote{
A very thorough discussion of the universal
{\it a priori\/} probability is given by \newcite{Li:93a}.}


\begin{figure*}[tb]
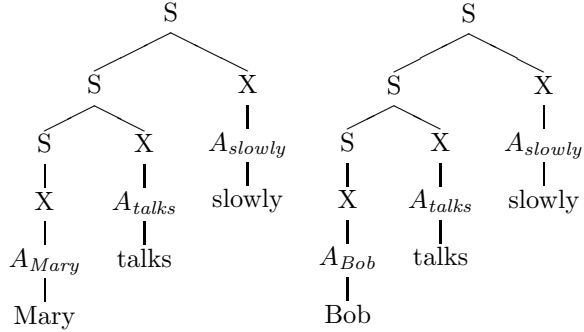


\leaf{Bob}
\branch{1}{$A_{\mbox{\scriptsize\it Bob}}$}
\branch{1}{X}
\branch{1}{S}
\leaf{talks}
\branch{1}{$A_{\mbox{\scriptsize\it talks}}$}
\branch{1}{X}
\branch{2}{S}
\leaf{slowly}
\branch{1}{$A_{\mbox{\scriptsize\it slowly}}$}
\branch{1}{X}
\branch{2}{S}

\leaf{Mary}
\branch{1}{$A_{\mbox{\scriptsize\it Mary}}$}
\branch{1}{X}
\branch{1}{S}
\leaf{talks}
\branch{1}{$A_{\mbox{\scriptsize\it talks}}$}
\branch{1}{X}
\branch{2}{S}
\leaf{slowly}
\branch{1}{$A_{\mbox{\scriptsize\it slowly}}$}
\branch{1}{X}
\branch{2}{S}

$$ \tree\tree $$

\caption{Initial Viterbi Parse} \label{fig:before}
\end{figure*}

\begin{figure*}[tb]
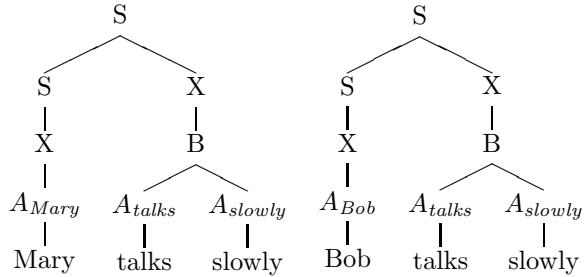


\leaf{Bob}
\branch{1}{\vbox{\vbox to2pt{} \hbox{$A_{\mbox{\scriptsize\it Bob}}$}}}
\branch{1}{X}
\branch{1}{S}
\leaf{talks}
\branch{1}{$A_{\mbox{\scriptsize\it talks}}$}
\leaf{slowly}
\branch{1}{$A_{\mbox{\scriptsize\it slowly}}$}
\branch{2}{B}
\branch{1}{X}
\branch{2}{S}

\leaf{Mary}
\branch{1}{\vbox{\vbox to2pt{} \hbox{$A_{\mbox{\scriptsize\it Mary}}$}}}
\branch{1}{X}
\branch{1}{S}
\leaf{talks}
\branch{1}{$A_{\mbox{\scriptsize\it talks}}$}
\leaf{slowly}
\branch{1}{$A_{\mbox{\scriptsize\it slowly}}$}
\branch{2}{B}
\branch{1}{X}
\branch{2}{S}

$$ \tree\tree $$

\caption{Predicted Viterbi Parse} \label{fig:after}
\end{figure*}

\section{Search Algorithm}

As described above, we take grammar induction to be the search
for the grammar $G'$ that optimizes the objective
function $p(O|G)p(G)$.  While this framework does not restrict us
to a particular grammar formalism, in our work we consider only
probabilistic context-free grammars.

We assume a simple greedy search strategy.  We maintain a single
hypothesis grammar which is initialized to a small, trivial grammar.
We then try to find a modification to the hypothesis grammar, such
as the addition of a grammar rule, that results in a grammar
with a higher score on the objective function.  When we find a superior
grammar, we make this the new hypothesis grammar.  We repeat
this process until we can no longer find a modification that improves
the current hypothesis grammar.

For our initial grammar, we choose a grammar that can generate
any string, to assure that the grammar can cover the training data.
The initial grammar is listed in Table \ref{tab:init}.
The sentential symbol $S$ expands to a sequence of $X$'s, where $X$
expands to every other nonterminal symbol in the grammar.
Initially, the set of nonterminal symbols consists of
a different nonterminal symbol expanding to each terminal symbol.

Notice that this grammar models a sentence as a sequence of independently
generated nonterminal symbols.  We maintain this property throughout
the search process, that is, for every symbol $A'$ that we add to the grammar,
we also add a rule $X \r A'$.  This assures that
the sentential symbol can expand to every symbol; otherwise, adding
a symbol will not affect the probabilities that the grammar assigns
to strings.

We use the term {\it move set\/} to describe the set of modifications
we consider to the current hypothesis grammar to hopefully produce
a superior grammar.  Our move set includes the following moves:
\begin{description}
\item[Move 1:] Create a rule of the form $A \r BC$
\item[Move 2:] Create a rule of the form $A \r B|C$
\end{description}
For any context-free grammar, it is possible to express a weakly equivalent
grammar using only rules of these forms.
As mentioned before, with each new symbol $A$ we also create a rule $X \r A$.

\subsection{Evaluating the Objective Function}

Consider the task of calculating the objective function
$p(O|G)p(G)$ for some grammar $G$.  Calculating $p(G) = 2^{-l(G)}$ is
inexpensive\footnote{Due to space limitations, we do not specify
our method for encoding grammars, \ie, how we calculate $l(G)$ for
a given $G$.  However, this will be described in the author's
forthcoming Ph.D. dissertation.};
however, calculating $p(O|G)$ requires a parsing of the
entire training data.  We cannot afford to parse the training
data for each grammar considered; indeed, to ever be practical
for data sets of millions of words,
it seems likely that we can only afford to parse the data once.

To achieve this goal, we employ several approximations.
First, notice that
we do not ever need to calculate the actual value of the objective function;
we need only to be able to distinguish when a move applied to the current
hypothesis grammar produces a grammar that has a higher score on the
objective function, that is, we need only to be able to calculate
the {\it difference\/} in the objective function resulting from a move.
This can be done efficiently if we can quickly approximate how the
probability of the training data changes when a move is applied.

To make this possible, we approximate the probability of the
training data $p(O|G)$ by the probability of the single most probable
parse, or {\it Viterbi\/} parse, of the training data.  Furthermore,
instead of recalculating the Viterbi parse of the training data from
scratch when a move is applied, we use heuristics to predict how a move
will change the Viterbi parse.
For example, consider the case where the training
data consists of the two sentences
$$ O = \{ \mbox{Bob talks slowly}, \mbox{Mary talks slowly} \} $$
In Figure \ref{fig:before}, we display the Viterbi parse of this
data under the initial hypothesis grammar used in our algorithm.

Now, let us consider the move of adding the rule
$$ B \r A_{\mbox{\scriptsize\it talks}} \; A_{\mbox{\scriptsize\it slowly}} $$
to the initial grammar
(as well as the concomitant rule $X \r B$).  A reasonable heuristic
for predicting how the Viterbi parse will change is to replace
adjacent $X$'s that expand to $A_{\mbox{\scriptsize\it talks}}$ and
$A_{\mbox{\scriptsize\it slowly}}$
respectively with a single $X$ that expands to $B$, as displayed
in Figure \ref{fig:after}.  This is the actual heuristic we use for moves
of the form $A \r BC$, and we have analogous heuristics for each move
in our move set.  By predicting the differences in the Viterbi parse resulting
from a move, we can quickly estimate the change in the probability of
the training data.

Notice that our predicted Viterbi parse can stray a great deal from
the actual Viterbi parse, as errors can accumulate as move after move
is applied.  To minimize these effects,
we process the training data incrementally.
Using our initial hypothesis grammar, we parse the first sentence
of the training data and search for the optimal
grammar over just that one sentence using the described
search framework.  We use the resulting grammar to parse the
second sentence, and then search for the optimal grammar over the
first two sentences using the last grammar as the starting point.
We repeat this process, parsing the next
sentence using the best grammar found on the previous sentences
and then searching for the best grammar taking into account this
new sentence, until the entire training corpus is covered.

Delaying the parsing of a sentence
until all of the previous sentences are processed should yield
more accurate Viterbi parses during the search process than if
we simply parse the whole corpus with the initial hypothesis grammar.
In addition, we still achieve the goal of parsing each sentence
but once.

\subsection{Parameter Training}

In this section, we describe how the parameters of our grammar,
the probabilities associated with each grammar rule, are set.
Ideally, in evaluating the objective function for a particular grammar
we should use its optimal parameter settings given the training data,
as this is the full score that the given grammar can achieve.
However, searching for optimal parameter values is extremely expensive
computationally.  Instead, we grossly approximate the optimal values by
deterministically setting parameters based on
the Viterbi parse of the training data parsed so far.
We rely on the post-pass, described later, to refine parameter values.

Referring to the rules in Table \ref{tab:init}, the parameter $\e$ is set to
an arbitrary small constant.  The values of the parameters $p(A)$ are
set to the (smoothed) frequency of the $X \r A$ reduction in the
Viterbi parse of the data seen so far.  The remaining symbols
are set to expand uniformly among their possible expansions.

\subsection{Constraining Moves}

Consider the move of creating a rule of the form $A \r BC$.
This corresponds to $k^3$ different specific rules that might be created,
where $k$ is the current number of symbols in the grammar.  As it is too
computationally expensive to consider each of these rules at every point in
the search, we use heuristics to constrain which moves are appraised.

For the left-hand side of a rule, we always create a new symbol.
This heuristic selects the optimal choice the vast majority of
the time; however, under this constraint the
moves described earlier in this section cannot yield arbitrary context-free
languages.  To partially address this, we add the move
\begin{description}
\item[Move 3:] Create a rule of the form $A \r AB | B$
\end{description}
With this iteration move, we can construct grammars that generate
arbitrary regular languages.  As yet, we have not implemented
moves that enable the construction of arbitrary context-free grammars;
this belongs to future work.

To constrain the symbols we consider on the right-hand side of a new rule,
we use what we call {\it triggers}.\footnote{This is not to be confused with
the use of the term {\it triggers\/} in dynamic language modeling.}
A {\it trigger\/} is a phenomenon in the Viterbi parse of a sentence that is
indicative that a particular move might lead to a better grammar.
For example, in Figure \ref{fig:before} the fact that
the symbols $A_{\mbox{\scriptsize\it talks}}$ and
$A_{\mbox{\scriptsize\it slowly}}$ occur
adjacently is indicative that it could be profitable to create
a rule $B \r A_{\mbox{\scriptsize\it talks}} A_{\mbox{\scriptsize\it slowly}}$.
We have developed a set of triggers for each move in our move set,
and only consider a specific move if it is triggered in the
sentence currently being parsed in the incremental processing.

\subsection{Post-Pass}

A conspicuous shortcoming in our search framework is
that the grammars in our search space are fairly unexpressive.
Firstly, recall that our grammars model a sentence as a sequence of
independently generated symbols; however, in language there is a large
dependence between adjacent constituents.  Furthermore, the only free
parameters
in our search are the parameters $p(A)$; all other symbols (except $S$) are
fixed to expand uniformly.  These choices were necessary to make the
search tractable.

To address this issue, we use an Inside-Outside algorithm post-pass.
Our methodology is derived from that
described by \newcite{Lari:90a}.  We create $n$ new nonterminal symbols
$\{X_1, \ldots, X_n\}$, and create all rules of the form:
$$ \begin{array}{cccll}
X_i & \r & X_j & X_k \hspace{0.3in} & i, j, k \in \{ 1, \ldots, n \} \\
X_i & \r & A & & i \in \{1, \ldots, n\}, \\
& & & & A \in N_{\mbox{\scriptsize\it old}} - \{S, X \}
\end{array} $$
$N_{\mbox{\scriptsize\it old}}$ denotes the set of nonterminal symbols
acquired in the initial grammar induction phase, and $X_1$ is taken to be the
new sentential symbol.  These new rules replace the first three rules
listed in Table \ref{tab:init}.
The parameters of these rules are initialized
randomly.  Using this grammar as the starting point, we
run the Inside-Outside algorithm on the training data until convergence.

In other words, instead of using the naive $S \r SX | X$ rule to attach
symbols together in parsing data, we now use
the $X_i$ rules and depend on the Inside-Outside algorithm to
train these randomly initialized rules intelligently.
This post-pass allows us to express dependencies between adjacent
symbols.  In addition, it allows us to train parameters that were
fixed during the initial grammar induction phase.


\section{Previous Work}

As mentioned, this work employs the Bayesian grammar induction framework
described by Solomonoff \shortcite{Solomonoff:60a,Solomonoff:64a}.
However, Solomonoff does not specify a concrete search
algorithm and only makes suggestions as to its nature.

Similar research includes work by Cook et al.\ (1976) and
Stolcke and Omohundro (1994).
This work also employs a heuristic
search within a Bayesian framework.  However, a different prior
probability on grammars is used, and the algorithms are only
efficient enough to be applied to small data sets.

The grammar induction algorithms most successful in language modeling
include the Inside-Outside algorithm \cite{Lari:90a,Lari:91a,Pereira:92a},
a special case of the Expectation-Maximization algorithm \cite{Dempster:77a},
and work by \newcite{McCandless:93a}.  In the latter work, McCandless uses a
heuristic search procedure similar to ours, but a very different
search criteria.  To our knowledge, neither algorithm has surpassed
the performance of $n$-gram models in a language modeling task
of substantial scale.


\begin{table*}[tb]

$$ \begin{tabular}{|l|c|c|c|c|} \hline
& best & entropy & entr.\ relative \\
& $n$ & (bits/word) & to $n$-gram \\ \hline
ideal grammar & & 2.30 & $-$6.5\% \\ \hline
our algorithm & 7 & 2.37 & $-$3.7\% \\ \hline
$n$-gram model & 4 & 2.46 & \\ \hline
Inside-Outside & 9 & 2.60 & $+$5.7\% \\ \hline
\end{tabular} $$
\caption{English-like artificial grammar \label{tab:eng}}

\end{table*}

\begin{table*}[tb]

$$ \begin{tabular}{|l|c|c|c|c|} \hline
& best & entropy & entr.\ relative \\
& $n$ & (bits/word) & to $n$-gram \\ \hline
ideal grammar & & 4.13 & $-$10.4\% \\ \hline
our algorithm & 9 & 4.44 & $-$3.7\% \\ \hline
$n$-gram model & 4 & 4.61 & \\ \hline
Inside-Outside & 9 & 4.64 & $+$0.7\% \\ \hline
\end{tabular} $$
\caption{Wall Street Journal-like artificial grammar \label{tab:wsj}}

\end{table*}

\begin{table*}[tb]

$$ \begin{tabular}{|l|c|c|c|c|} \hline
& best & entropy & entr.\ relative \\
& $n$ & (bits/word) & to $n$-gram \\ \hline
$n$-gram model & 6 & 3.01 & \\ \hline
our algorithm & 7 & 3.15 & $+$4.7\% \\ \hline
Inside-Outside & 7 & 3.93 & $+$30.6\% \\ \hline
\end{tabular} $$
\caption{English sentence part-of-speech sequences \label{tab:pos}}

\end{table*}

\begin{table*}[tb]
$$ \begin{tabular}{|l|c|r|r|r|} \hline
WSJ & $n$ & entropy & no. & time \\
artif. & & (bits/word) & params & (sec) \\ \hline
$n$-gram & 3 & 4.61 & 15000 & 50 \\ \hline
IO & 9 & 4.64 & 2000 & 30000 \\ \hline
first pass & & & 800 & 1000 \\ \hline
post-pass & 5 & 4.60 & 4000 & 5000 \\ \hline
\end{tabular} $$
\caption{Parameters and Training Time \label{tab:param}}
\end{table*}

\section{Results}

To evaluate our algorithm, we compare the performance of our algorithm
to that of $n$-gram models and the Inside-Outside algorithm.

For $n$-gram models, we tried $n = 1, \ldots, 10$ for each domain.
For smoothing a particular $n$-gram model, we took a linear combination
of all lower order $n$-gram models.  In particular, we follow standard
practice \cite{Jelinek:80a,Bahl:83a,Brown:90a} and
take the smoothed $i$-gram probability to be a linear combination
of the $i$-gram frequency in the training data and the smoothed $(i-1)$-gram
probability, that is,
\begin{eqnarray*}
\lefteqn{p(w_0 | W = w_{i-1} \cdots w_{-1}) =} \\
& & \l_{i,c(W)} \frac{c(Ww_0)}{c(W)} + \\
& & (1 - \l_{i,c(W)}) p(w_0 | w_{i-2} \cdots w_{-1})
\end{eqnarray*}
where $c(W)$ denotes the count of the word sequence $W$ in the training data.
The smoothing parameters $\l_{i,c}$ are trained through
the Forward-Backward algorithm \cite{Baum:67a} on held-out data.
Parameters $\l_{i,c}$ are tied together for similar $c$ to prevent
data sparsity.

For the Inside-Outside algorithm, we follow the methodology described
by Lari and Young.  For a given $n$, we create a probabilistic
context-free grammar consisting
of all Chomsky normal form rules over the $n$ nonterminal
symbols $\{X_1, \ldots X_n\}$ and the given terminal symbols, that is, all
rules
$$ \begin{array}{cccll}
X_i & \r & X_j & X_k \hspace{0.3in} & i, j, k \in \{ 1, \ldots, n \} \\
X_i & \r & a & & i \in \{1, \ldots, n\}, a \in T
\end{array} $$
where $T$ denotes the set of terminal symbols in the domain.  All parameters
are initialized randomly.  From this starting point, the Inside-Outside
algorithm is run until convergence.

For smoothing, we combine the expansion distribution of each
symbol with a uniform distribution, that is, we take the smoothed parameter
$p_s(A \r \a)$ to be
$$ p_s(A \r \a) = (1 - \l) p_u(A \r \a) + \l \frac{1}{n^3 + n|T|} $$
where $p_u(A \r \a)$ denotes the unsmoothed parameter.  The value
$n^3 + n|T|$ is the number of different ways a symbol expands under
the Lari and Young methodology.  The parameter $\l$ is trained through the
Inside-Outside algorithm on held-out data.  This smoothing is
also performed on the Inside-Outside post-pass of our algorithm.
For each domain, we tried $n = 3, \ldots, 10$.

Because of the computational demands of our algorithm, it is
currently impractical to apply it to large vocabulary or
large training set problems.  However, we present the results of
our algorithm in three medium-sized domains.  In each case,
we use 4500 sentences for training, with 500 of these sentences held out
for smoothing.  We test on 500 sentences, and measure performance
by the entropy of the test data.

In the first two domains, we created the training and test data
artificially so as to have an ideal grammar in hand to benchmark results.
In particular, we used a probabilistic grammar to generate
the data.  In the first domain, we created this grammar by hand;
the grammar was a small English-like probabilistic
context-free grammar consisting
of roughly 10 nonterminal symbols, 20 terminal symbols, and 30 rules.
In the second domain, we derived the grammar from manually
parsed text.  From a million words of parsed Wall Street Journal
data from the Penn treebank, we extracted the 20 most
frequently occurring symbols, and the 10 most frequently occurring rules
expanding each of these symbols.  For each symbol that occurs
on the right-hand side of a rule but which was not one of
the most frequent 20 symbols, we create a rule that expands
that symbol to a unique terminal symbol.  After removing unreachable rules,
this yields a grammar of roughly 30 nonterminals, 120 terminals, and 160 rules.
Parameters are set to reflect the frequency of the
corresponding rule in the parsed corpus.

For the third domain, we took English text and reduced the size of
the vocabulary by mapping each word to its part-of-speech tag.  We used
tagged Wall Street Journal text from the Penn treebank, which has
a tag set size of about fifty.

In Tables \ref{tab:eng}--\ref{tab:pos}, we summarize our results.
The {\it ideal grammar\/} denotes the grammar used to generate
the training and test data.  For each algorithm, we list
the best performance achieved over all $n$ tried,
and the {\it best $n$} column states which value realized
this performance.

We achieve a moderate but significant improvement in performance over
$n$-gram models and the Inside-Outside algorithm in the first two domains,
while in the part-of-speech domain we are outperformed by $n$-gram models
but we vastly outperform the Inside-Outside algorithm.

In Table \ref{tab:param}, we display a sample of the number of parameters and
execution time (on a Decstation 5000/33) associated with each algorithm.
We choose $n$ to yield approximately equivalent performance for each
algorithm.  The {\it first pass\/} row refers to the main grammar induction
phase of our algorithm, and the {\it post-pass\/} row refers to
the Inside-Outside post-pass.

Notice that our algorithm produces a significantly more compact model
than the $n$-gram model, while running significantly faster than the
Inside-Outside algorithm even though we use an Inside-Outside post-pass.
Part of this discrepancy is due to the fact that we require a smaller
number of new nonterminal symbols to achieve equivalent performance, but
we have also found that our post-pass converges more quickly even given
the same number of nonterminal symbols.



\section{Discussion}

Our algorithm consistently outperformed the Inside-Outside algorithm
in these experiments.  While we partially attribute this difference to
using a Bayesian instead of maximum-likelihood objective function, we
believe that part of this difference results from a more effective
search strategy.  In particular, though both algorithms employ
a greedy hill-climbing strategy, our algorithm gains an advantage
by being able to add new rules to the grammar.

In the Inside-Outside algorithm, the gradient descent search
discovers the ``nearest'' local minimum in the search landscape to the
initial grammar.  If there are $k$ rules in the grammar and thus $k$
parameters, then the search takes place in a fixed $k$-dimensional
space ${\Bbb R}^k$.  In our algorithm, it is possible to expand
the hypothesis grammar, thus increasing the
dimensionality of the parameter space that is being searched.
An apparent local minimum in the space ${\Bbb R}^k$ may no longer be a local
minimum in the space ${\Bbb R}^{k+1}$; the extra dimension may provide
a pathway for further improvement of the hypothesis grammar.
Hence, our algorithm should be less prone to
suboptimal local minima than the Inside-Outside algorithm.

Outperforming $n$-gram models in the first two domains demonstrates
that our algorithm is able to take advantage of the grammatical
structure present in data.
However, the superiority of $n$-gram models in the part-of-speech
domain indicates that to be competitive in modeling naturally-occurring
data, it is necessary to model collocational information accurately.  We need
to modify our algorithm to more aggressively model $n$-gram information.


\section{Conclusion}

This research represents a step forward in the quest for
developing grammar-based language models for natural language.
We induce models that, while being substantially more compact,
outperform $n$-gram language models in medium-sized domains.
The algorithm runs essentially in time and space linear in
the size of the training data, so larger domains are within our
reach.

However, we feel the largest contribution of this work does not lie in
the actual algorithm specified, but rather in its indication of the
potential of the induction framework described by Solomonoff
in 1964.  We have implemented only a subset of the moves that
we have developed, and inspection
of our results gives reason to believe that these additional moves may
significantly improve the performance of our algorithm.

Solomonoff's induction framework is not restricted to
probabilistic context-free grammars.  After completing
the implementation of our move set, we plan to explore
the modeling of context-sensitive phenomena.  This work demonstrates
that Solomonoff's elegant framework deserves much further consideration.

\section*{Acknowledgements}

We are indebted to Stuart Shieber for his suggestions and guidance,
as well as his invaluable comments on earlier drafts
of this paper.  This material is based on work supported by the
National Science Foundation under Grant Number IRI-9350192 to
Stuart M. Shieber.



\begin{thebibliography}{}

\bibitem[\protect\citename{Angluin and Smith}1983]{Angluin:83a}
D.~Angluin and C.H. Smith.
\newblock 1983.
\newblock Inductive inference: theory and methods.
\newblock {\em ACM Computing Surveys}, 15:237--269.

\bibitem[\protect\citename{Bahl \bgroup et al.\egroup }1978]{Bahl:78a}
L.R. Bahl, J.K. Baker, P.S. Cohen, F.~Jelinek, B.L. Lewis, and R.L. Mercer.
\newblock 1978.
\newblock Recognition of a continuously read natural corpus.
\newblock In {\em Proceedings of the IEEE International Conference on
  Acoustics, Speech and Signal Processing}, pages 422--424, Tulsa, Oklahoma,
  April.

\bibitem[\protect\citename{Bahl \bgroup et al.\egroup }1983]{Bahl:83a}
Lalit~R. Bahl, Frederick Jelinek, and Robert~L. Mercer.
\newblock 1983.
\newblock A maximum likelihood approach to continuous speech recognition.
\newblock {\em IEEE Transactions on Pattern Analysis and Machine Intelligence},
  PAMI-5(2):179--190, March.

\bibitem[\protect\citename{Baker}1975]{Baker:75a}
J.K. Baker.
\newblock 1975.
\newblock The {DRAGON} system -- an overview.
\newblock {\em IEEE Transactions on Acoustics, Speech and Signal Processing},
  23:24--29, February.

\bibitem[\protect\citename{Baker}1979]{Baker:79b}
J.K. Baker.
\newblock 1979.
\newblock Trainable grammars for speech recognition.
\newblock In {\em Proceedings of the Spring Conference of the Acoustical
  Society of America}, pages 547--550, Boston, MA, June.

\bibitem[\protect\citename{Baum and Eagon}1967]{Baum:67a}
L.E. Baum and J.A. Eagon.
\newblock 1967.
\newblock An inequality with application to statistical estimation for
  probabilistic functions of {M}arkov processes and to a model for ecology.
\newblock {\em Bulletin of the American Mathematicians Society}, 73:360--363.

\bibitem[\protect\citename{Brown \bgroup et al.\egroup }1992]{Brown:90a}
Peter~F. Brown, Vincent~J. DellaPietra, Peter~V. deSouza, Jennifer~C. Lai, and
  Robert~L. Mercer.
\newblock 1992.
\newblock Class-based n-gram models of natural language.
\newblock {\em Computational Linguistics}, 18(4):467--479, December.

\bibitem[\protect\citename{Dempster \bgroup et al.\egroup }1977]{Dempster:77a}
A.P. Dempster, N.M. Laird, and D.B. Rubin.
\newblock 1977.
\newblock Maximum likelihood from incomplete data via the {EM} algorithm.
\newblock {\em Journal of the Royal Statistical Society}, 39(B):1--38.

\bibitem[\protect\citename{Jelinek and Mercer}1980]{Jelinek:80a}
Frederick Jelinek and Robert~L. Mercer.
\newblock 1980.
\newblock Interpolated estimation of {M}arkov source parameters from sparse
  data.
\newblock In {\em Proceedings of the Workshop on Pattern Recognition in
  Practice}, Amsterdam, The Netherlands: North-Holland, May.

\bibitem[\protect\citename{Kernighan \bgroup et al.\egroup
  }1990]{Kernighan:90a}
M.D. Kernighan, K.W. Church, and W.A. Gale.
\newblock 1990.
\newblock A spelling correction program based on a noisy channel model.
\newblock In {\em Proceedings of the Thirteenth International Conference on
  Computational Linguistics}, pages 205--210.

\bibitem[\protect\citename{Lari and Young}1990]{Lari:90a}
K.~Lari and S.J. Young.
\newblock 1990.
\newblock The estimation of stochastic context-free grammars using the
  inside-outside algorithm.
\newblock {\em Computer Speech and Language}, 4:35--56.

\bibitem[\protect\citename{Lari and Young}1991]{Lari:91a}
K.~Lari and S.J. Young.
\newblock 1991.
\newblock Applications of stochastic context-free grammars using the
  inside-outside algorithm.
\newblock {\em Computer Speech and Language}, 5:237--257.

\bibitem[\protect\citename{Li and Vit\'anyi}1993]{Li:93a}
Ming Li and Paul Vit\'anyi.
\newblock 1993.
\newblock {\em An Introduction to {K}olmogorov Complexity and its
  Applications}.
\newblock Springer-Verlag.

\bibitem[\protect\citename{McCandless and Glass}1993]{McCandless:93a}
Michael~K. McCandless and James~R. Glass.
\newblock 1993.
\newblock Empirical acquisition of word and phrase classes in the {ATIS}
  domain.
\newblock In {\em Third European Conference on Speech Communication and
  Technology}, Berlin, Germany, September.

\bibitem[\protect\citename{Pereira and Schabes}1992]{Pereira:92a}
Fernando Pereira and Yves Schabes.
\newblock 1992.
\newblock Inside-outside reestimation from partially bracket corpora.
\newblock In {\em Proceedings of the 30th Annual Meeting of the ACL}, pages
  128--135, Newark, Delaware.

\bibitem[\protect\citename{Resnik}1992]{Resnik:92a}
P.~Resnik.
\newblock 1992.
\newblock Probabilistic tree-adjoining grammar as a framework for statistical
  natural language processing.
\newblock In {\em Proceedings of the 14th International Conference on
  Computational Linguistics}.

\bibitem[\protect\citename{Rissanen}1978]{Rissanen:78a}
J.~Rissanen.
\newblock 1978.
\newblock Modeling by the shortest data description.
\newblock {\em Automatica}, 14:465--471.

\bibitem[\protect\citename{Schabes}1992]{Schabes:92a}
Y.~Schabes.
\newblock 1992.
\newblock Stochastic lexicalized tree-adjoining grammars.
\newblock In {\em Proceedings of the 14th International Conference on
  Computational Linguistics}.

\bibitem[\protect\citename{Shannon}1951]{Shannon:51a}
C.E. Shannon.
\newblock 1951.
\newblock Prediction and entropy of printed {E}nglish.
\newblock {\em Bell Systems Technical Journal}, 30:50--64, January.

\bibitem[\protect\citename{Solomonoff}1960]{Solomonoff:60a}
R.J. Solomonoff.
\newblock 1960.
\newblock A preliminary report on a general theory of inductive inference.
\newblock Technical Report ZTB-138, Zator Company, Cambridge, MA, November.

\bibitem[\protect\citename{Solomonoff}1964]{Solomonoff:64a}
R.J. Solomonoff.
\newblock 1964.
\newblock A formal theory of inductive inference.
\newblock {\em Information and Control}, 7:1--22, 224--254, March, June.

\bibitem[\protect\citename{Srihari and Baltus}1992]{Srihari:92a}
Rohini Srihari and Charlotte Baltus.
\newblock 1992.
\newblock Combining statistical and syntactic methods in recognizing
  handwritten sentences.
\newblock In {\em AAAI Symposium: Probabilistic Approaches to Natural
  Language}, pages 121--127.

\end{thebibliography}
\end{document}